\documentclass[5p,twocolumn,authoryear]{elsarticle}

\usepackage[T1]{fontenc}
\usepackage[utf8]{inputenc}
\usepackage{amsmath}
\usepackage[charter]{mathdesign}  
\usepackage{graphicx}
\usepackage{booktabs}
\usepackage{multirow}
\usepackage{array}
\usepackage[font=footnotesize,labelfont=bf,labelsep=period]{caption}
\usepackage{fancyhdr}
\usepackage{orcidlink}
\usepackage[dvipsnames]{xcolor}
\usepackage{microtype}
\usepackage{hyperref}
\definecolor{scooToxBlue}{HTML}{3776AB}
\hypersetup{colorlinks=true}
\AtBeginDocument{\hypersetup{linkcolor=scooToxBlue,citecolor=scooToxBlue,urlcolor=scooToxBlue}}

\newcommand{\doi}[1]{\href{https://doi.org/#1}{\nolinkurl{https://doi.org/#1}}}

\newcommand{\shorttitle}{Kinematic signatures of impairment in e-scooter riders}
\newcommand{\shortauthors}{R.R. Pai et al.}

\makeatletter
\def\ps@pprintTitle{%
  \let\@oddhead\@empty
  \let\@evenhead\@empty
  \def\@oddfoot{\footnotesize\itshape Preprint -- under peer review\hfill\today}%
  \let\@evenfoot\@oddfoot}
\makeatother

\newcommand{\figplaceholder}[2]{%
  \fbox{\parbox[c][\dimexpr#1*35/100\relax][c]{\dimexpr#1*95/100\relax}{%
    \centering\small\ttfamily Missing figure: \detokenize{#2}}}}
\newcommand{\insertfig}[2][\linewidth]{%
  \IfFileExists{#2.pdf}{\includegraphics[width=#1]{#2.pdf}}{%
  \IfFileExists{#2.png}{\includegraphics[width=#1]{#2.png}}{%
  \IfFileExists{#2.jpg}{\includegraphics[width=#1]{#2.jpg}}{%
  \figplaceholder{#1}{#2}}}}}

\newcommand{\pv}[2]{$#1\times10^{#2}$}
\newcommand{\pvb}[2]{{\boldmath$#1\times10^{#2}$}}

\begin{document}

\begin{frontmatter}

\title{Kinematic signatures of impairment: Detecting alcohol intoxication in e-scooter riders using sensor data and machine learning}

\author[1,2]{Rahul Rajendra Pai\,\orcidlink{0000-0002-1516-6930}\corref{cor1}}
\ead{rahul.pai@chalmers.se}
\author[1]{Marco Dozza\,\orcidlink{0000-0002-6544-4281}}
\author[1]{Alexander Rasch\,\orcidlink{0000-0001-6868-8364}}
\author[1]{Ali Mohammadi\,\orcidlink{0000-0001-9285-5994}}
\author[2]{Marco Capuccini\,\orcidlink{0000-0002-4851-759X}}

\cortext[cor1]{Corresponding author.}

\affiliation[1]{organization={Chalmers University of Technology, Division of Vehicle Safety, Department of Mechanical Engineering},
  city={Gothenburg},
  country={Sweden}}
\affiliation[2]{organization={Voi Technology AB},
  city={Stockholm},
  country={Sweden}}

\begin{abstract}
Alcohol intoxication is a leading contributor to fatal and severe-injured e-scooterist crashes. Current countermeasures, such as temporal restrictions or pre-ride cognitive screening, cannot continuously assess an e-scooterist's physical motor control or impairment in real time. We conducted a controlled experiment in which 25 participants rode an instrumented e-scooter through a test track while sober and at two targeted blood alcohol concentration levels (0.05\% and 0.08\%). The e-scooter was instrumented with a six-axis inertial measurement unit (IMU), and throttle and brake lever position sensors, all sampled at 100~Hz. Two complementary signal features were computed: normalised permutation entropy, which quantifies temporal complexity, and standard deviation, which quantifies signal amplitude. Repeated measures correlation identified seven kinematic features (all IMU and throttle signals) whose entropy decreased ($p < 0.001$) while standard deviation increased ($p < 0.01$) with increasing intoxication, indicating that intoxicated riders shift from continuous, low-amplitude micro-corrections to fewer, high-amplitude reactive corrections. An entropy based multi-class logistic regression classifier, evaluated through leave-one-participant-out cross-validation, achieved 85\% overall accuracy and a weighted one-vs-rest area under the receiver operating characteristic curve (AuROC) of 0.94, with a sober-vs-high AuROC of 1.00. Steering rate and lateral acceleration were the most important predictive features, indicating that alcohol induces a distinct collapse in lateral equilibrium during riding. Ultimately, these results demonstrate that onboard kinematic sensing combined with entropy-based signal analysis can reliably distinguish sober from intoxicated e-scooter riding, providing a foundation for automatic intoxication detection systems that preserve mobility for sober riders.
\end{abstract}


\begin{keyword}
Micromobility safety \sep e-scooter anomaly detection \sep drunk riding identification \sep permutation entropy \sep vehicle kinematics \sep rider behavioural analysis
\end{keyword}

\end{frontmatter}

\section{Introduction}

The rapid adoption of electric scooters (e-scooters) for urban transport has introduced safety challenges that are only beginning to be understood. In Sweden, a recent in-depth analysis of all fatal micromobility crashes (2016--2024) found that 44\% of e-scooter fatalities involved alcohol-intoxicated riders, a prevalence substantially higher than for conventional bicyclists (13\%) or e-cyclists (27\%) \citep{pai2026three}. This pattern is not unique to fatal outcomes. Non-fatal injury data report alcohol involvement in 12\% of e-scooter crashes in Washington, D.C. \citep{cicchino2021severity}, 31.6\% in a German emergency department where intoxication was associated with a 27-fold increase in major injury risk \citep{hartz2025characteristics}, and up to 46.5\% in Helsinki prior to the introduction of nighttime restrictions \citep{dibaj2024exploration}. Together, these findings identify alcohol intoxication as one of the most critical modifiable risk factors in e-scooter safety.

Despite the clear association between intoxication and crash risk, regulatory and technological responses remain limited. Sweden currently imposes no legal blood alcohol concentration (BAC) limit for e-scooterists, unlike neighbouring Nordic countries that have established thresholds ranging from 0.02\% in Norway to 0.05\% in Denmark, Finland, and Iceland \citep{radet2025regler,statens2025trafikkregler,stjornarrad2025rafmagn,traficom2025sahkoiset}. Even where such limits exist, enforcement through roadside testing of micromobility users presents practical challenges that make compliance difficult to monitor at scale. Several rental operators have introduced in-app reaction tests---screen-based responsiveness checks---as proxies for riding readiness. Meanwhile, some municipalities have imposed blanket nighttime riding bans, which address the temporal concentration of intoxicated crashes but penalise sober riders and constrain the utility of e-scooters as a transport mode.

Controlled studies have established a mechanistic basis for the observed crash risk. \citet{zube2022escooter} demonstrated in a test-track experiment that alcohol intoxication progressively degraded e-scooter riding performance. Experimental cycling studies have documented similar impairments: \citet{hartung2015regarding} showed that motoric degradation becomes measurable at BAC levels of 0.8\textperthousand, with complete loss of sober-equivalent performance by 1.4\textperthousand, while \citet{andersson2023bicycling} found that stability decreased with intoxication on a treadmill-based cycling task, with standard deviation (SD) of roll rate identified as an indicator of instability. \citet{anderses2024importance} extended these findings by showing that individual characteristics modulate the relationship between alcohol and cycling performance. Together, these results suggest that the kinematic signatures of intoxicated riding are sufficiently distinct to be detectable by onboard sensing, a hypothesis that, to date, has not been tested for e-scooters.

An onboard, continuous, and automatic detection system based on vehicle kinematics would represent a paradigm shift from the current approaches. Rather than restricting access for all users based on time of day or relying on a single pre-ride screen test that is easily circumvented, such a system would monitor the e-scooterist's control behaviour throughout the ride and intervene only when impairment is detected. This approach would preserve full mobility for sober users while targeting the specific population responsible for the disproportionate share of serious and fatal crashes. The concept is analogous to driver drowsiness detection systems now standard in passenger vehicles, which infer impairment from steering-wheel input patterns.

The objective of this study was to determine whether alcohol intoxication in e-scooterists can be reliably detected from onboard kinematic signals and characterise the nature of the kinematic change induced by intoxication. We hypothesised that the temporal complexity of kinematic signals, quantified through permutation entropy (PE), decreases with increasing intoxication, while the amplitude of kinematic signals, quantified through standard deviation, increases---together reflecting a shift from continuous, low-amplitude micro-corrections to fewer, high-amplitude reactive corrections. To test this hypothesis, we conducted a controlled experiment in which participants rode an instrumented e-scooter through a test track at progressive intoxication levels and developed a classification framework to identify the kinematic signatures of alcohol impairment during e-scooter riding.

\section{Method}

\subsection{Participants and ethics}

Thirty-three individuals (24 male, 9 female) participated in the experiment. Eight participants were excluded due to data loss during the experiment, leaving 25 participants (19 males, 6 females; mean age $= 26.5 \pm 4.6$ years) in the final sample. The included participants had a mean weight of 81.36~kg ($\pm$~17.58~kg), a mean height of 179.08~cm ($\pm$~8.51~cm), and a mean body mass index (BMI) of 25.42 ($\pm$~5.70). The demographics of the excluded participants were comparable to those of the retained participants, and no other exclusion criteria were applied.

Inclusion criteria required participants to be over 20 years of age, able to communicate in English, capable of riding a bicycle. Exclusion criteria included pregnancy or breastfeeding, any ongoing somatic or psychiatric disease, pathological injuries limiting mobility, a history of serious traffic crashes, current medication use, or drug therapies known to interact with alcohol.

Alcohol consumption habits were screened to ensure participants were regular but moderate consumers of alcohol. Participants were required to have consumed at least two alcoholic beverages per month in the preceding six months, without exceeding 14 units per week for females or 21 units per week for males (one unit defined as 33~cl of 5.2\% beer, 4~cl of 40\% liquor, or 15~cl of 13\% wine). The Alcohol Use Disorders Identification Test (AUDIT) was administered, with exclusion thresholds set at scores of $> 14$. To control for metabolic variability, participants were instructed to have an early breakfast but to abstain from lunch before the data collection sessions, which commenced at 14:00.

The study protocol was approved by the Swedish Ethical Review Authority (Etikpr\"ovningsmyndigheten) (Ref.~2025-04833), and all participants provided written informed consent prior to participation. A medical advisor was on call throughout all experimental sessions. Helmet use was mandatory, and supplementary protective gears (wrist, elbow and knee protection devices) were provided. After the experiment, participants were monitored until their BAC showed a downward trend and were discharged only when accompanied by a sober individual or provided with a taxi ride home.

\subsection{Apparatus and environment}

\subsubsection{Instrumented e-scooter}

The experiment used a Segway Apex D110L commercial-grade e-scooter shown in Fig.~\ref{fig:setup}. The cruising speed was limited to 7~km/h, matching the speed restriction applied in geofenced pedestrian zones within Swedish cities. The speed restriction was also motivated by participant safety and to accommodate the spatial constraints of the indoor test track.

The e-scooter was instrumented with the following sensors (Fig.~\ref{fig:setup}A): a 6-axis Inertial Measurement Unit (IMU), mounted on the steering column, measuring triaxial acceleration (lateral, $a_x$; vertical, $a_y$; longitudinal, $a_z$) and triaxial angular velocity (pitch rate, $r_x$; steering rate, $r_y$; roll rate, $r_z$); a wheel-speed sensor deriving speed from the e-scooter's motor; and position sensors embedded in the brake and throttle levers, measuring the continuous position of the left brake lever (brake$_l$), right brake lever (brake$_r$), and throttle. All signals were logged synchronously at 100~Hz using an inbuilt data logger \citep{pai2025understanding}.

\subsubsection{Test track}

\begin{figure*}[t]
  \centering
  \insertfig[0.9\textwidth]{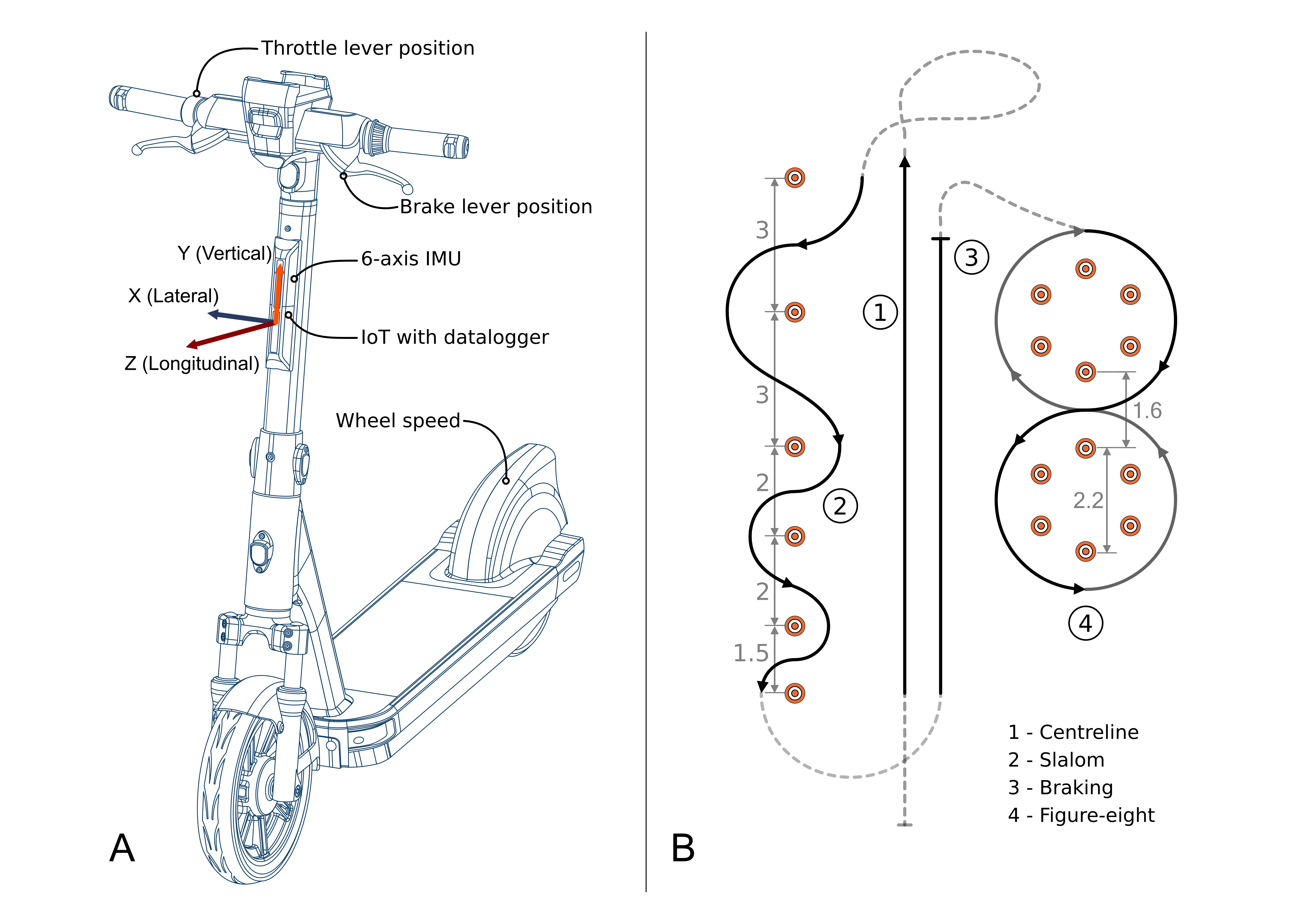}
  \caption{Instrumentation on the vehicle (Panel A) and the layout of the MicroLab test track (Panel B).}
  \label{fig:setup}
\end{figure*}

The experiment was conducted at the MicroLab test facility at Chalmers University of Technology, an indoor facility with a concrete floor surface, controlling for weather conditions and variable road friction. The test track (Fig.~\ref{fig:setup}B) comprised four sequential manoeuvres designed to elicit distinct kinematic demands: (1) a straight-line path (centreline) to establish baseline cruising speed and line tracking ability; (2) a slalom section with six cones, where cone spacing progressively narrowed from 3~m (cones 1--3), to 2~m (cones 3--5), to 1.5~m (cones 5--6); (3) a braking zone, in which the rider returned along the straight-line path and executed a complete stop with the front wheel on a ground marking; and (4) a figure-eight pattern consisting of two adjacent circles of 2.2~m diameter with edges 1.6~m apart, completed twice per trial.

\subsection{Experimental procedure}

\subsubsection{Familiarisation phase}

Prior to data collection, participants were given open-ended time to practise riding the e-scooter while sober to mitigate confounding due to learning effects. Participants practised until they felt comfortable with the vehicle and the track layout.

\subsubsection{Experiment protocol}

Each participant completed the test protocol under three conditions: sober (baseline), low intoxication (target BAC $= 0.05\%$), and high intoxication (target BAC $= 0.08\%$). Within each intoxication condition, participants completed the trial twice. The conditions were administered sequentially in a single session: sober trials were completed first, followed by the administration of the first alcohol dose, a 15-minute absorption period, and the low-intoxication trials. A second alcohol dose was then administered, followed by another 15-minute absorption period, and the high-intoxication trials.

The final dataset comprised 141 valid trials (45 Sober, 50 Low, 46 High). Nine trials across eight of the 25 participants were excluded due to occasional data logger memory exhaustion during recording.

\subsubsection{Intoxication protocol}

Participants chose vodka (40\% alcohol by volume), rum (35--40\%), or whiskey (35\%) at the start of the experiment and maintained the same beverage throughout. The beverage was consumed neat or mixed with a non-alcoholic beverage of the participant's choice (e.g., water, orange juice, or cola). The volume of alcoholic beverage required to reach the target BAC was calculated for each individual using a modified Widmark formula \citep{dozza2026mintox} (see \ref{app:alcohol}).

BAC was measured using a Dr\"ager Alcotest 6000, a law-enforcement-grade breathalyser, which measures breath alcohol concentration (BrAC) and applies internal conversion factors to report BAC. A BAC reading was recorded immediately prior to each individual trial, yielding two distinct BAC measurements per intoxication level. While the target BAC levels were 0.05\% and 0.08\%, no adjustments were made if participants undershot or overshot the targets; participants proceeded with the protocol as planned. The achieved mean BAC distributions were: Low $= 0.035 \pm 0.008\%$; High $= 0.065 \pm 0.008\%$.

\subsection{Data analysis}

\subsubsection{Pre-processing}

Trial initiation was defined as the first instance at which the vehicle speed exceeded 0~km/h. Trial termination was determined through synchronised external optical tracking to establish the time required for the rider to complete the track layout. To isolate human kinematic inputs from high-frequency chassis vibration, a 64th-order finite-impulse-response (FIR) low-pass filter with a Hamming window \citep{oppenheim2010discrete} and a cutoff frequency of 10~Hz was applied bidirectionally to all six IMU channels achieving a steep cutoff. The cutoff frequency was chosen because voluntary human motor control occurs predominantly below 6~Hz \citep{winter2009biomechanics}; a cutoff at 10~Hz therefore preserves the full bandwidth of rider-generated signals without attenuation while suppressing mechanical noise. The throttle and brake lever position signals were not filtered, as these represent discrete rider control inputs rather than high-frequency mechanical noise. Missing data points (fewer than 10\% for any given trial) were imputed using linear interpolation before low-pass filtering.

\subsubsection{Feature extraction and within-subject centring}

Normalised permutation entropy \citep{bandt2002permutation} was computed for each of the nine signals ($a_x$, $a_y$, $a_z$, $r_x$, $r_y$, $r_z$, throttle, brake$_l$, brake$_r$). PE quantifies the complexity of a time series by measuring the diversity of ordinal patterns in the signal; values close to 1 indicate high irregularity, while values close to 0 indicate regularity or predictability. Two parameters govern the ordinal pattern construction: the embedding dimension $m$, which defines the length of subsequences used to form a pattern, and the time delay $\tau$, which specifies the number of samples skipped between consecutive points in each subsequence. The embedding dimension was set to $m = 5$, yielding $m! = 120$ possible ordinal patterns. The time delay was derived from the Nyquist relationship \citep{oppenheim2010discrete} applied to the filtered signal:
\begin{equation}
  \tau = \left[\frac{f_s}{2\cdot f_c}\right]
  \label{eq:tau}
\end{equation}
where $f_s = 100$~Hz is the sampling rate and $f_c = 10$~Hz is the low-pass cutoff frequency, yielding $\tau = 5$ samples. PE values were normalised by $\log_2(m!)$.

In addition to PE, the standard deviation of each signal was computed per trial as a measure of signal amplitude (i.e., the magnitude of deviations from the signal mean). While PE captures the temporal organisation of the rider's corrections, SD quantifies their magnitude; together, the two measures characterise complementary aspects of motor control degradation. The use of SD as a measure of kinematic variability under intoxication is established in the cycling literature, where \citet{andersson2023bicycling} identified the SD of yaw and roll rate increased systematically with BrAC.

In the modelling step, to account for individual differences in baseline riding behaviour, within-subject centring (WSC) was applied. The inductive WSC approach prevents information leakage during cross-validation. For each trial $i$, the centred feature vector was computed by subtracting the mean of all other trials from the same participant, excluding trial $i$:
\begin{equation}
  X_{\mathrm{wsc},i} = X_i - \frac{1}{\left|S_p \setminus \{i\}\right|} \sum_{j \in S_p \setminus \{i\}} X_j
  \label{eq:wsc}
\end{equation}

\subsubsection{Feature selection}

To identify signals associated with intoxication, repeated measures correlations \citep{bland1995a,bland1995b} were computed between each PE and SD value and the ordinal intoxication level (Sober $= 0$, Low $= 1$, High $= 2$) \citep{bakdash2017repeated}. Two complementary one-sided tests were applied: For PE, the alternative hypothesis was $H_1\colon \rho < 0$, testing the expectation that PE decreases---that is, signals become more regular and predictable---with increasing alcohol dose; for SD, the alternative hypothesis was $H_1\colon \rho > 0$, testing the expectation that SD increases---that is, signal amplitude increases---with intoxication. Because each hypothesis was tested across nine sensor channels, the Holm--Bonferroni sequential correction \citep{holm1979simple} was applied separately to the PE and SD $p$-value families to control the family-wise error rate at $\alpha = 0.05$. After correction, all IMU and throttle signals remained significantly associated with intoxication in both the PE and SD analyses (PE: $p < 0.001$; SD: $p < 0.01$), while neither brake signal reached the significance threshold in either of the analyses ($\alpha = 0.05$). Consequently, the seven significant features were retained for both the PE-based and SD-based classification.

\subsubsection{Classification model and baselines}

Four classification models were evaluated to assess intoxication detection performance and to compare the discriminative value of temporal complexity (PE) against signal amplitude (SD).

A multi-class logistic regression (LR) model trained on the seven significant WSC-PE features, with L2 regularisation ($C = 1.0$) served as the primary classifier, selected for its computational efficiency and interpretability. The model was optimised using the limited-memory Broyden--Fletcher--Goldfarb--Shanno (L-BFGS) quasi-Newton solver \citep{nocedal2006numerical}.

Two WSC-PE based and one WSC-SD based baseline models were evaluated for comparison. The first was a non-machine-learning sum-entropy threshold model, which collapsed the seven WSC-PE values into a single scalar score $s_i$ per trial:
\begin{equation}
  s_i = \sum_{j=1}^{7} X_{\mathrm{wsc},i,j}
  \label{eq:sum}
\end{equation}
Within each cross-validation fold, class-conditional means of the training scores ($\mu_H$, $\mu_L$, $\mu_S$) were computed, and two decision thresholds were placed at their respective midpoints: $t_1 = (\mu_H + \mu_L)/2$ and $t_2 = (\mu_L + \mu_S)/2$. A test trial was classified as High if $s_i < t_1$, Low if $t_1 \le s_i < t_2$, and Sober otherwise.

The second baseline was a support vector machine (SVM) with a radial basis function (RBF) kernel, also trained on the seven WSC-PE features, optimised using the deterministic LIBSVM solver with default hyperparameters from the scikit-learn library ($C = 1.0$, $\gamma = 1 / (n_{\mathrm{features}} \cdot \mathrm{Var}(X))$). Probability scores for the SVM were obtained by applying a softmax transformation to the one-vs-rest decision function values. The SVM was implemented to verify whether nonlinear decision boundaries improve classification performance relative to the LR model.

The third baseline was an LR classifier identical in architecture and hyperparameters to the primary model but trained on WSC-SD features instead. The model was implemented to quantify the relative discriminative value of temporal complexity versus signal amplitude.

\subsubsection{Cross-validation and evaluation metrics}

Model performance was evaluated using leave-one-participant-out (LOPO) cross-validation. Across 25 folds, all trials from a single participant were held out as the test set while models were trained on the remaining 24 participants. Within each fold, the seven selected WSC features were standardised to zero mean and unit variance using a scaler fitted exclusively on the training partition. All evaluation metrics were computed on the concatenated out-of-fold predictions across all 25 LOPO folds. Performance was assessed using overall classification accuracy and macro-averaged per-class precision, recall, and F1-score. The area under the receiver operating characteristic curve (AuROC) was calculated in a one-vs-rest (OvR) framework. Additionally, a sober-vs-high AuROC was computed by subsetting predictions to only the High and Sober trials, using the predicted probability of the High class. For the sum-entropy baseline, class-specific proximity scores were derived from the scalar score $s_i$: the Sober proximity score used the raw $s_i$, the High proximity score used the inverted score $-s_i$, and the Low proximity score was defined as the negative absolute distance from the Low class mean, i.e.\ $-|s_i - \mu_L|$.

\subsubsection{Feature importance}

To assess which signals contributed most to intoxication classification, the model coefficients ($\beta$) of the PE-based logistic regression model were examined. The coefficient matrix (intoxication level $\times$ features) was averaged across all 25 LOPO folds to account for cross-validation variance. The overall importance of each feature $j$ was calculated as the sum of absolute mean coefficients across classes $c$:
\begin{equation}
  \mathrm{Importance}_j = \sum_{c} \left|\bar{\beta}_{c,j}\right|
  \label{eq:importance}
\end{equation}

All data processing, feature extraction, and statistical modelling were performed in Python 3.10. Signal filtering was implemented using SciPy \citep{virtanen2020scipy}, permutation entropy was computed using the AntroPy library \citep{vallat2021antropy}, and all classification models and CV procedures were implemented using scikit-learn \citep{pedregosa2011scikit}. The analysis code and anonymised experimental data are publicly available at: \url{https://github.com/voi-oss/impairment-detection}.

\section{Results}

\subsection{Feature selection}

\begin{figure*}[t]
    \centering
    \insertfig[0.9\textwidth]{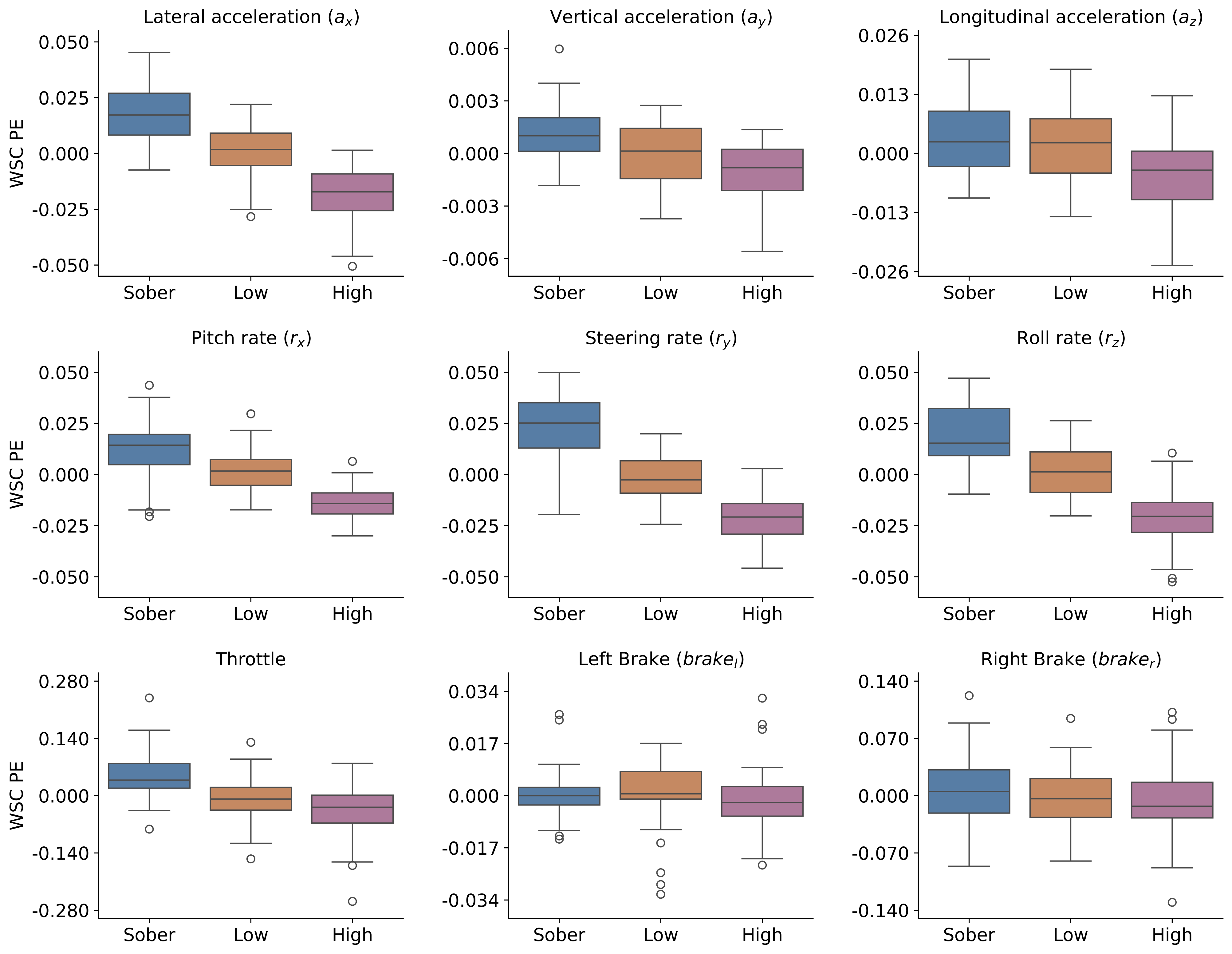}
    \caption{Within-subject centred (WSC) permutation entropy (PE) distributions for all nine evaluated signals stratified by intoxication condition (sober, low, high).}
    \label{fig:pe}
\end{figure*}

Repeated measures correlations between PE values and intoxication level revealed that all IMU signals and the throttle signal exhibited significant negative correlations with intoxication ($p < 0.001$) (Table~\ref{tab:rmcorr}). The steering rate ($r_y$) showed the strongest association ($p < 0.001$), followed by roll rate ($r_z$; $p < 0.001$) and lateral acceleration ($a_x$; $p < 0.001$). The distributions of WSC-PE values across the three intoxication levels are shown in Fig.~\ref{fig:pe}. For the feature selection, transductive WSC was computed by subtracting each participant's mean across all conditions to visualise the within-subject effect; the classification models used inductive centring. For the selected features, a consistent downward trend in PE is visible from the Sober to High condition.

\begin{figure*}[t]
  \centering
  \insertfig[0.9\textwidth]{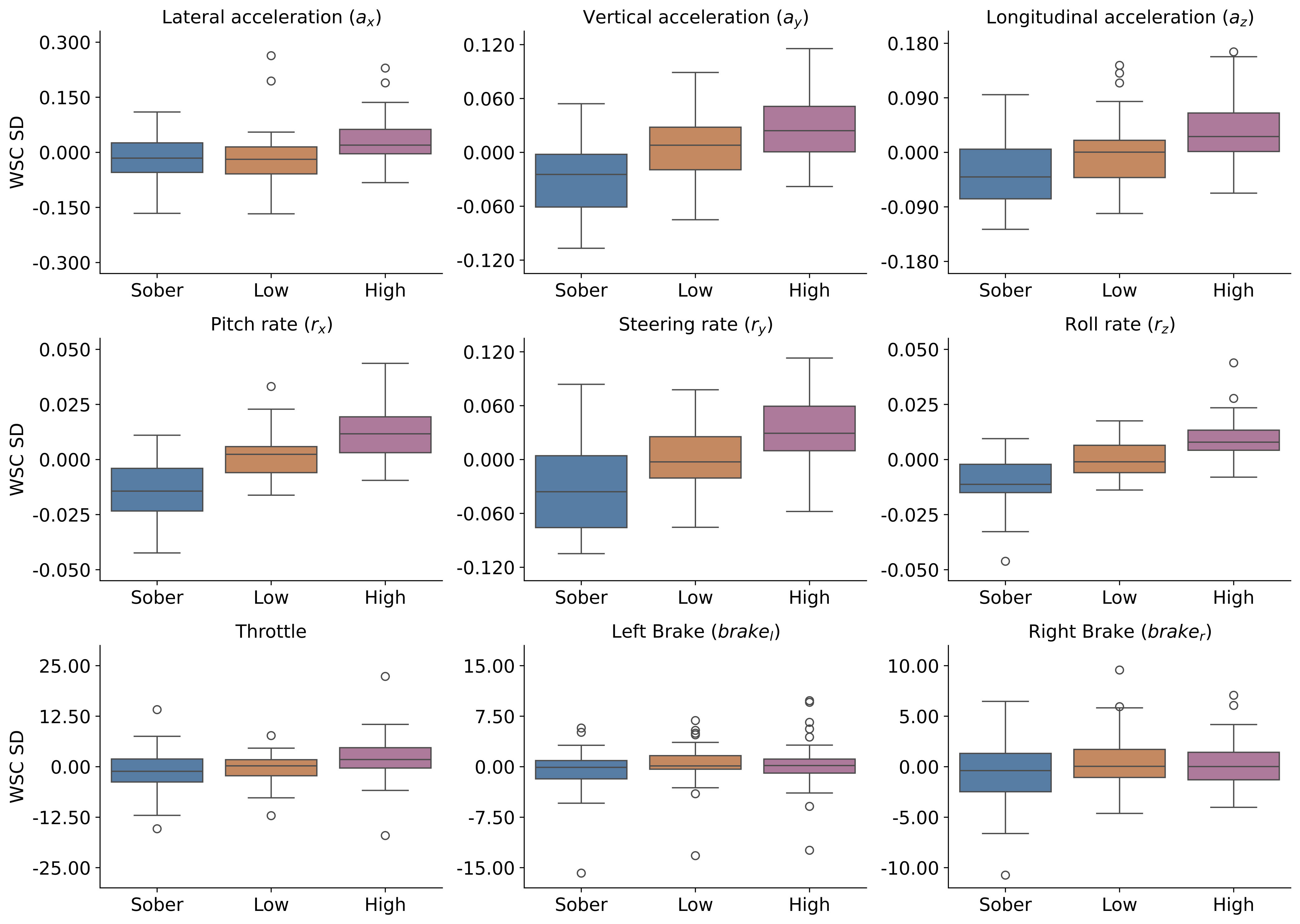}
  \caption{Within-subject centred (WSC) standard deviation (SD) distributions for all nine evaluated signals stratified by intoxication condition (sober, low, high).}
  \label{fig:sd}
\end{figure*}

Repeated measures correlations between SD values and intoxication level revealed that all IMU signals and the throttle signal exhibited significant positive correlations with alcohol dose ($p < 0.01$) (Table~\ref{tab:rmcorr}). The angular velocity signals ($r_x$, $r_y$ and $r_z$) showed the strongest SD associations, all reaching $p < 0.001$. Among the acceleration signals, vertical ($a_y$) and longitudinal ($a_z$) acceleration also reached $p < 0.001$, whereas lateral acceleration ($a_x$) and throttle were significant at $p < 0.01$ but did not reach $p < 0.001$. The distributions of transductive WSC-SD values across the three intoxication levels are shown in Fig.~\ref{fig:sd}. The brake signals did not reach significance and were excluded from all classification models.

\begin{table}[t]
  \caption{Repeated measures correlations between features and intoxication level. PE was tested with $H_1\colon \rho < 0$ (complexity decreases); SD was tested with $H_1\colon \rho > 0$ (amplitude increases).}
  \label{tab:rmcorr}
  \footnotesize
  \begin{tabular*}{\columnwidth}{@{\extracolsep{\fill}}lll@{}}
    \toprule
    Feature & PE $p$-value & SD $p$-value \\
    \midrule
    $r_y$ (steering rate)             & \pvb{2.50}{-28} & \pvb{1.75}{-10} \\
    $r_z$ (roll rate)                 & \pvb{1.79}{-24} & \pvb{7.81}{-13} \\
    $a_x$ (lateral acceleration)      & \pvb{5.75}{-21} & \pv{8.68}{-3}   \\
    $r_x$ (pitch rate)                & \pvb{1.01}{-16} & \pvb{1.07}{-14} \\
    throttle                          & \pvb{1.92}{-8}  & \pv{9.07}{-3}   \\
    $a_y$ (vertical acceleration)     & \pvb{6.47}{-6}  & \pvb{1.98}{-8}  \\
    $a_z$ (longitudinal acceleration) & \pvb{8.11}{-5}  & \pvb{3.51}{-6}  \\
    brake$_l$                         & 0.089           & 0.065           \\
    brake$_r$                         & 0.089           & 0.065           \\
    \bottomrule
  \end{tabular*}

  \smallskip
  {\scriptsize\textit{Note:} Bold font indicates statistical significance ($p < 0.001$).}
\end{table}

\subsection{Classification performance}

The classification results for all four models are presented in Table~\ref{tab:perf} and Table~\ref{tab:auroc}. The permutation entropy based logistic regression model achieved an overall accuracy of 85\%, with per-class F1-scores of 0.89 (Sober), 0.78 (Low), and 0.88 (High). The SVM achieved marginally higher accuracy (86\%) with comparable per-class performance. Both machine learning models substantially outperformed the sum-entropy threshold baseline (74\% accuracy).

In terms of discriminative ability (Table~\ref{tab:auroc}), the PE-based logistic regression model achieved a weighted one-vs-rest AuROC of 0.94 and a sober-vs-high AuROC of 1.00. The SVM performed comparably (weighted AuROC $= 0.94$; sober-vs-high AuROC $= 0.99$). The sum-entropy threshold baseline, despite its simplicity, achieved a sober-vs-high AuROC of 0.99.

The SD-based logistic regression achieved lower performance than its PE-based counterpart despite identical model architecture: 61\% overall accuracy, a weighted AuROC of 0.79, and a sober-vs-high AuROC of 0.90. The Low intoxication class was the most difficult to discriminate across all models, consistent with it representing an intermediate state.

\begin{table*}[t]
  \caption{Three-class classification performance (LOPO CV, 25 folds).}
  \label{tab:perf}
  \footnotesize
  \begin{tabular*}{\textwidth}{@{\extracolsep{\fill}}lcccccccccc@{}}
    \toprule
    & & \multicolumn{3}{c}{High} & \multicolumn{3}{c}{Low} & \multicolumn{3}{c}{Sober} \\
    \cmidrule(lr){3-5}\cmidrule(lr){6-8}\cmidrule(l){9-11}
    Model & Accuracy & Precision & Recall & F1 score & Precision & Recall & F1 score & Precision & Recall & F1 score \\
    \midrule
    Sum entropy                & 0.74 & 0.78 & 0.78 & 0.78 & 0.62 & 0.68 & 0.65 & 0.85 & 0.76 & 0.80 \\
    WSC-PE logistic regression & 0.85 & 0.87 & 0.89 & 0.88 & 0.81 & 0.76 & 0.78 & 0.87 & 0.91 & 0.89 \\
    Support vector machine     & 0.86 & 0.86 & 0.96 & 0.91 & 0.84 & 0.74 & 0.79 & 0.87 & 0.89 & 0.88 \\
    WSC-SD logistic regression & 0.61 & 0.62 & 0.63 & 0.62 & 0.53 & 0.50 & 0.52 & 0.68 & 0.71 & 0.70 \\
    \bottomrule
  \end{tabular*}
\end{table*}

\begin{table*}[t]
  \caption{AuROC scores under LOPO CV.}
  \label{tab:auroc}
  \footnotesize
  \begin{tabular*}{\textwidth}{@{\extracolsep{\fill}}lccccc@{}}
    \toprule
    Model & Sober-vs-rest & Low-vs-rest & High-vs-rest & Weighted & Sober-vs-high \\
    \midrule
    Sum entropy                & 0.9475 & 0.7923 & 0.9297 & 0.8867 & 0.9937 \\
    WSC-PE logistic regression & 0.9664 & 0.8912 & 0.9682 & 0.9403 & 0.9995 \\
    Support vector machine     & 0.9667 & 0.9002 & 0.9625 & 0.9417 & 0.9884 \\
    WSC-SD logistic regression & 0.8782 & 0.6936 & 0.8206 & 0.7940 & 0.9024 \\
    \bottomrule
  \end{tabular*}

  \smallskip
  {\scriptsize\textit{Note:} The `Weighted' column represents the prevalence-weighted average of the three one-vs-rest (Sober-vs-rest, Low-vs-rest, High-vs-rest) classification scores.}
\end{table*}

\subsection{Feature importance}

\begin{figure}[t]
  \centering
  \insertfig[\columnwidth]{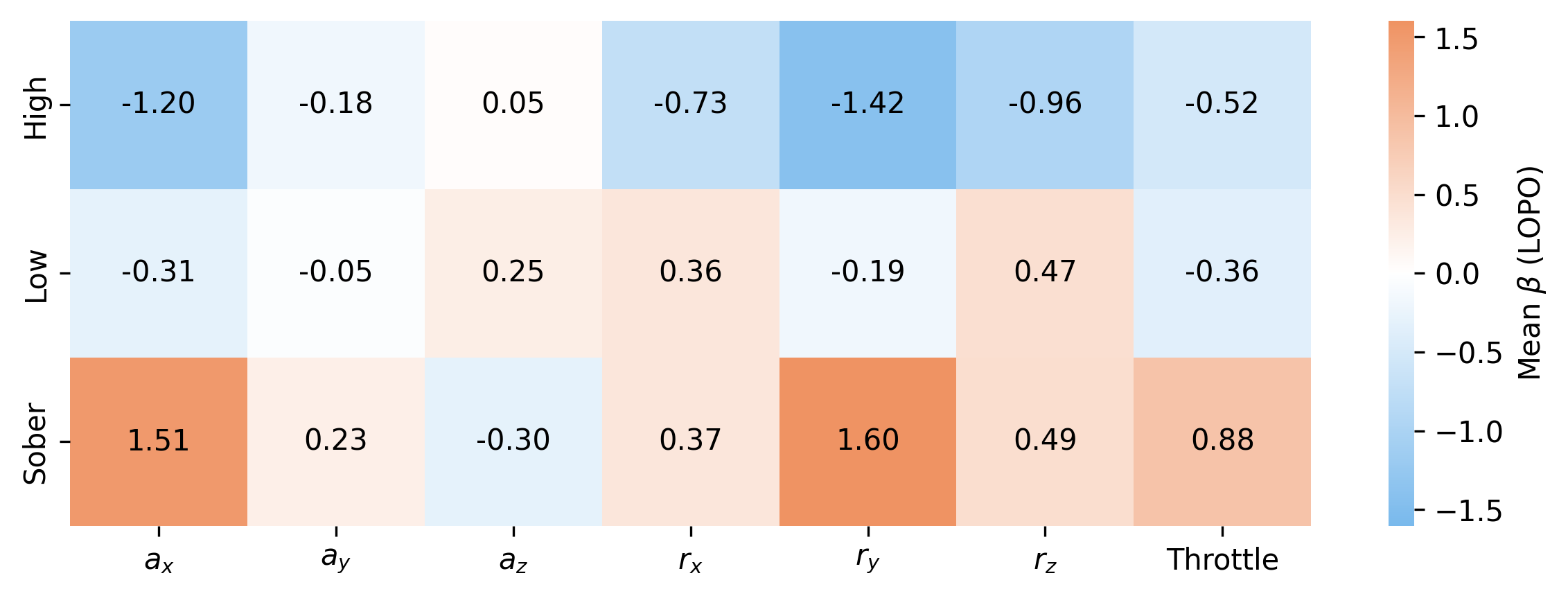}
  \caption{Mean standardised entropy-based logistic regression coefficients for the seven retained features across the three intoxication classes.}
  \label{fig:coef}
\end{figure}

The mean logistic regression coefficients of the WSC-PE-based model averaged across all 25 LOPO folds are presented in Fig.~\ref{fig:coef} as a heatmap. The overall feature importance ranking (Table~\ref{tab:importance}) identified steering rate ($r_y$; $\Sigma|\beta| = 3.21$) and lateral acceleration ($a_x$; $\Sigma|\beta| = 3.03$) as the dominant predictive features, followed by roll rate ($r_z$; 1.93) and throttle (1.76). The vertical ($a_y$) and longitudinal ($a_z$) acceleration components contributed least to the classification.

\begin{table}[t]
  \caption{Feature importance ranking based on the summed absolute mean WSC-PE-based LR coefficients across all LOPO folds.}
  \label{tab:importance}
  \footnotesize
  \begin{tabular*}{\columnwidth}{@{\extracolsep{\fill}}lc@{}}
    \toprule
    Feature & $\Sigma\,|\beta|$ \\
    \midrule
    $r_y$ (steering rate)             & 3.21 \\
    $a_x$ (lateral acceleration)      & 3.03 \\
    $r_z$ (roll rate)                 & 1.93 \\
    throttle                          & 1.76 \\
    $r_x$ (pitch rate)                & 1.47 \\
    $a_z$ (longitudinal acceleration) & 0.60 \\
    $a_y$ (vertical acceleration)     & 0.47 \\
    \bottomrule
  \end{tabular*}
\end{table}

\section{Discussion}

This study demonstrated that alcohol intoxication in e-scooterists can be reliably detected from onboard kinematic signals using permutation entropy and a logistic regression classifier. The primary model achieved 85\% overall accuracy and, critically, near-perfect discrimination between sober and highly intoxicated riding (AuROC $= 1.00$). These results establish that the kinematic signatures of intoxicated e-scooter riding are sufficiently distinct for automated detection.

\subsection{Rider impairment characterisation during intoxicated riding}

The central finding that PE systematically decreases with increasing intoxication reveals a fundamental change in the rider's neuromuscular control strategy \citep{lipsitz1992loss}. During sober riding, a rider continuously executes high-frequency micro-adjustments to maintain balance, track the intended path, and respond to environmental perturbations. These ongoing corrections produce a kinematic signal with diverse temporal patterns and, consequently, high entropy. Under the influence of alcohol, the capacity for these continuous micro-adjustments is progressively degraded \citep{creaser2009effects,dong2024drinking,koelega1995alcohol,moskowitz2000review,tzambazis2000alcohol}. It is important to note that the decrease in PE does not imply smaller steering movements; rather, it indicates that the temporal ordering of movements became more predictable. The simultaneous increase in SD establishes that when intoxicated, riders make larger corrections. The intoxicated rider therefore shifts from a strategy of continuous, low-amplitude corrections to one characterised by high-amplitude reactive corrections. The signal becomes more regular not because control improves, but because the adaptive, moment-to-moment variability that characterises skilled motor control is suppressed \citep{parks2002effects}. This interpretation aligns with the motor control literature, which shows that reduced movement complexity has been associated with impaired neuromuscular function across a range of tasks \citep{stergiou2011human}. The finding is also consistent with experimental cycling studies, which have documented progressive motoric degradation with increasing BAC \citep{hartung2015regarding}, and with the test-track e-scooter study of \citet{zube2022escooter}, which observed degraded riding performance under intoxication.

\subsection{Directional importance of kinematic features}

The feature importance analysis revealed a clear directional hierarchy: signals capturing lateral dynamics---steering rate and lateral acceleration---were substantially more important for detecting intoxication than those capturing vertical or longitudinal dynamics. This pattern is interpretable in the context of e-scooter riding mechanics. Single-track vehicles such as e-scooters and bicycles behave as inverted pendulums \citep{kooijman2011bicycle} and are inherently unstable in the roll axis. Lateral balance and steering control are the primary active tasks during e-scooter riding; the rider must continuously adjust the handlebar to maintain the vehicle's lateral stability, particularly at the low speeds used in this experiment. The dominance of lateral signals is consistent with the bicycle intoxication experiments of \citet{andersson2023bicycling}, who identified roll rate as the most adequate indicator of cycling instability under alcohol, and \citet{anderses2024importance}, who confirmed the sensitivity of lateral stability measures to intoxication across individuals with different characteristics. At a more fundamental level, postural control research has shown that alcohol intoxication produces greater instability in the lateral direction than in the anteroposterior direction, likely due to slowed nerve impulse transmission to muscles controlling lateral movements \citep{modig2012blood,woollacott1983effects}. The present findings extend this observation from standing balance to active vehicle control during e-scooter riding.

Vertical acceleration, by contrast, is largely governed by the road surface and vehicle suspension rather than rider input, which explains its minimal contribution to intoxication detection. Longitudinal acceleration is primarily determined by the throttle and braking inputs rather than continuous balance corrections, and its low importance suggests that intoxicated riders can maintain broadly similar speed profiles to sober riders even as their lateral control deteriorates.

The dominance of steering rate as the most important feature parallels findings in automobile driver impairment detection, where steering-wheel reversal rate and steering entropy have been identified as sensitive indicators of drowsiness and distraction \citep{irwin2017effects,lee2010assessing}.

\subsection{Model performance and practical implications}

The PE-based logistic regression classifier achieved performance comparable to the more complex SVM (85\% vs 86\% accuracy; AuROC 0.94 vs 0.94), suggesting that the relationship between WSC-PE features and intoxication is approximately linear in the feature space and that added model complexity does not meaningfully improve detection. The practical significance of this finding is considerable: a linear model is more interpretable, requires fewer computational resources, and is more readily deployed on the embedded hardware of an e-scooter than a kernel-based model.

Even the simple sum-entropy threshold model---a non-machine-learning approach that collapses seven features into a single scalar---achieved 74\% accuracy and a sober-vs-high AuROC of 0.99. This result highlights the strength of the entropy-based signal: a substantial portion of the discriminative information is captured by the aggregate entropy decrease, and the multi-feature logistic regression primarily improves discrimination of the intermediate (Low) intoxication class.

While the preceding comparisons evaluated different modelling strategies applied to the same WSC-PE features, a separate question is whether signal amplitude alone can achieve comparable discrimination. The WSC-SD based classifier confirms that the sensitivity of amplitude-based measures to intoxication, previously demonstrated for cycling by \citet{andersson2023bicycling}, extends to e-scooter riding, achieving moderate sober-vs-high discrimination (AuROC $= 0.90$) but substantially lower overall accuracy (61\%) and weighted AuROC (0.79) than the PE-based classifier. This performance gap demonstrates that while signal amplitude carries some discriminative information, it is the temporal organisation of motor control that enables reliable three-class classification. Complementing PE with amplitude-based features may yield further improvements, though the present results indicate that temporal complexity alone carries the majority of the discriminative information.

It should be noted that the achieved BAC levels (Low: $0.035 \pm 0.008\%$; High: $0.065 \pm 0.008\%$) were systematically below the target values of 0.05\% and 0.08\%, likely reflecting individual variability in alcohol absorption kinetics and the 15-minute absorption period. The undershoot has two implications for interpreting the results. First, the classification performance reported here was achieved at BAC levels lower than intended, suggesting that the kinematic signatures of intoxication are detectable at relatively modest alcohol doses---a finding that strengthens rather than weakens the case for onboard detection. Second, the results may not be directly extrapolated to the higher BAC levels; classifier performance at 0.08\% or higher observed among the fatal crashes \citep{pai2026three} remains an empirical question that future studies should address.

From an application perspective, the near-perfect sober-vs-high AuROC (1.00 for PE-based logistic regression) is the most practically relevant result. A real-world deployment would likely operate as a binary sober-vs-impaired detector rather than a three-class system, and the very high discriminative ability between these extreme classes suggests that such a system could operate with very low false-positive rates. Deploying such a system on rental fleets is technically straightforward, as these vehicles already carry the necessary sensor hardware and connectivity infrastructure. However, as recent epidemiological evidence has shown that the majority of alcohol-related e-scooter fatalities involve privately owned vehicles \citep{pai2026three}, the greatest safety benefit would require extending detection capabilities to private e-scooters, which is a challenging proposition. The wide variation in private e-scooter designs and sensor configurations complicates the development of a universal detection system, and, more fundamentally, e-scooterists who purchase a private e-scooter have little incentive to adopt the system and pay the extra cost.

It should also be noted that the current PE-based LR model evaluates trial-level data; transitioning to real-time impairment monitoring requires an online processing architecture. This may be achieved by computing the PE features over a continuous sliding window or using sequential anomaly detection architectures \citep{capuccini2025testing}.

\subsection{Limitations and future research}

The primary limitation of this study is that the experiment was conducted at a cruising speed of 7~km/h---the speed limit applied in geofenced pedestrian zones in Swedish cities---in an indoor environment, whereas real-world e-scooter riding typically occurs at speeds up to 20--25~km/h in outdoor traffic. However, the controlled indoor environment warrants safety, eliminates confounds from uneven surfaces, traffic interactions, and weather, all of which would introduce variability in a real-world deployment. Future studies should investigate detection performance at operational speeds, ideally in naturalistic outdoor environments \citep{pai2025understanding} or simulators \citep{li2025laboratory}.

The intoxication protocol administered alcohol in a fixed ascending order (sober $\rightarrow$ low $\rightarrow$ high) within a single session, without counterbalancing the order of conditions across participants. The protocol was necessitated by the pharmacokinetics of alcohol intoxication: achieving a lower BAC after a higher one within a single session would require impractically long washout periods. A sober control group repeating the same protocol without alcohol would have isolated any practice or fatigue effects from intoxication but was not feasible within the scope and funding of the present study. While the familiarisation period ensures participants reach a performance plateau and WSC mitigates order effects, the design does not fully control for all potential fatigue or practice confounds.

Eight of the 33 enrolled participants (24\%) were excluded due to data loss caused by intermittent memory exhaustion of the onboard data logger. The experimental protocol required simultaneous coordination of breath alcohol measurement, timed dosing intervals, sensor initialisation, and external optical tracking across multiple trials per participant, all managed by a small research team within a single-session design. Under these operational demands, the logger memory limitation---identified during early sessions---could not be fully resolved without interrupting the ongoing data collection campaign. The failures occurred during signal recording and were unrelated to participant behaviour or intoxication condition; the demographics of the excluded participants were comparable to those of the retained sample. Nevertheless, the resulting attrition rate is high for a controlled experiment of this size and reduces statistical power. Future implementations should incorporate redundant onboard storage or real-time data integrity checks to prevent similar losses.

The brake lever signals did not contribute to classification, which may reflect the specific track design (a single braking event per trial) rather than a fundamental insensitivity of braking behaviour to intoxication. Track designs with more frequent braking demands may reveal additional discriminative information.

Finally, this study focused exclusively on quantifying the kinematic outcomes of impairment rather than the underlying neuromotor causes. Future studies should incorporate gaze-tracking and full-body motion-capture systems to decouple the rider's motor inputs from the vehicle's mechanics. Quantifying variables such as centre-of-gravity shifts and visual scanning behaviour will help distinguish between delayed feedforward planning and impaired feedback execution, providing deeper insight into the neuromotor deficits induced by alcohol.

\section{Conclusion}

While shared e-scooters are a widely adopted and generally safe mode of urban transit, their rapid integration has altered micro-mobility risk profiles, often outpacing the development of vehicle-specific safety countermeasures. This study establishes that alcohol impairment during e-scooter operation manifests not as a generalised degradation of vehicle control, but as a distinct, quantifiable collapse of lateral dynamic equilibrium. Intoxicated riders lose the adaptive high-frequency micro-corrections required to continuously stabilise the vehicle, shifting instead to fewer, high-amplitude reactive corrections---a change concentrated in steering and roll kinematics that is both measurable from onboard sensors and sufficiently distinct for automated detection.

Consequently, this research provides the foundation for continuous motor monitoring as a complement to existing countermeasures. Current approaches such as pre-ride screening and nighttime bans address intoxicated riding either before the ride begins or after a crash occurs; none monitors the rider's control during the ride itself. The demonstrated feasibility of extracting signal complexity from standard onboard telemetry shows that this gap can be closed using hardware already present on commercial e-scooter fleets. In the near term, kinematic monitoring can augment existing measures by targeting the specific riding behaviour associated with serious and fatal crashes; as detection accuracy improves, it may reduce reliance on blanket restrictions that penalise sober riders. This framework preserves the accessibility of shared e-scooters while concentrating intervention where the risk is greatest.

\section*{Data availability}

The anonymised experimental dataset and all analysis code used to generate the results reported in this study are publicly available at \url{https://github.com/voi-oss/impairment-detection} and \url{https://doi.org/10.6084/m9.figshare.34012614}.

\section*{Acknowledgements}

This work was carried out as part of the MicroTox project, funded by VINNOVA (Sweden's innovation agency) and Drive Sweden under grant number 2025-00431. The data collection was additionally supported by the MinToX project, funded by the Area of Advance Transport at Chalmers University of Technology and St\"od till MinTOX, funded by Trafikverket under grant number TRV2024/106354. This work was also partially supported by the Wallenberg AI, Autonomous Systems and Software Program (WASP) funded by the Knut and Alice Wallenberg Foundation.

The authors would like to express their sincere gratitude to the participants who generously contributed their time to this study. We extend our thanks to Niyathi Kini for her assistance during the experimental trials. We are also grateful to Andrea de Bejczy at the University of Gothenburg for providing medical support and oversight throughout the study. Furthermore, we thank Rikard Karlsson and Tord Hansson for enabling access to the Eventhallen at Chalmers, as well as Henrik H\"orlin, Peter B\"ackgren, and Kristina Henricson Briggs for facilitating access to the Tracks facilities at Chalmers. We thank Rahman Amandius from Voi for coordinating the industry partnership and facilitating the administrative requirements necessary for the collaboration.

\section*{CRediT authorship contribution statement}

\textbf{Rahul Rajendra Pai:} Methodology, Formal analysis, Validation, Visualization, Writing -- original draft, Writing -- review \& editing.
\textbf{Marco Dozza:} Conceptualization, Methodology, Supervision, Project administration, Funding acquisition, Writing -- review \& editing.
\textbf{Alexander Rasch:} Investigation, Data curation, Writing -- review \& editing.
\textbf{Ali Mohammadi:} Investigation, Data curation, Writing -- review \& editing.
\textbf{Marco Capuccini:} Methodology, Software, Supervision, Formal analysis, Validation, Funding acquisition, Writing -- review \& editing.

\section*{Declaration of generative AI and AI-assisted technologies in the manuscript preparation process}

During the preparation of this work, the author(s) used large language models to correct grammar and review language. After using this tool/service, the author(s) reviewed and edited the content as needed and take(s) full responsibility for the content of the article.

\section*{Declaration of competing interests}

At the time this research was conducted and submitted, Marco Capuccini reports a relationship with Voi Technology AB that includes: employment. Rahul Rajendra Pai reports a relationship with Voi Technology AB that includes: non-financial support. All other authors declare that they have no known competing financial interests or personal relationships that could have appeared to influence the work reported in this paper.

\appendix

\section{Alcohol volume estimation}
\label{app:alcohol}

The required beverage volume ($V_{\mathrm{beverage}}$) in millilitres was calculated as:
\begin{equation}
  V_{\mathrm{beverage}} =
  \frac{\dfrac{\left[10 \times \left(\mathrm{Target} - \mathrm{previous\_BAC}\right) + \left(\beta \times T\right)\right] \times V_d}{1 - R_d}}
       {\rho_{\mathrm{ethanol}} \times C_{\mathrm{alcohol}}}
  \label{eq:volume}
\end{equation}
where:
\begin{itemize}
  \item Target is the desired intoxication level (0.05 or 0.08).
  \item previous\_BAC is the participant's most recent measured alcohol concentration (set to 0 for the first dose).
  \item $\beta$ is the average alcohol elimination rate, set to 0.15 per hour.
  \item $T$ is the time elapsed for absorption, set to 0.25 hours (15 minutes).
  \item $R_d$ is the resorption deficit, which is the proportion of alcohol lost during gastrointestinal absorption, estimated at 15\% \citep{zube2022escooter}.
  \item $\rho_{\mathrm{ethanol}}$ is the density of ethanol (0.789~g/mL).
  \item $C_{\mathrm{alcohol}}$ is the alcohol content of the chosen beverage (0.40 or 0.35).
  \item $V_d$ is the volume of distribution, calculated as the participant's weight (kg) multiplied by the Widmark factor ($r$).
\end{itemize}

The Widmark factor ($r$) was calculated using the anthropometric equations of \citet{seidl2000calculation}:
\begin{align}
  r_{\mathrm{male}}   &= 0.31608 - 0.004821 \times \mathrm{weight} + 0.004632 \times \mathrm{height} \\
  r_{\mathrm{female}} &= 0.31223 - 0.006446 \times \mathrm{weight} + 0.004466 \times \mathrm{height}
\end{align}
where weight is in kilograms and height in centimetres.

\bibliographystyle{elsarticle-harv}
\bibliography{references}

\end{document}